\documentclass{article}
\usepackage{spconf,amsmath,amssymb,graphicx,epsfig,mathptmx,times,multirow,mathrsfs,subcaption}
\usepackage{hyperref}
\usepackage{cite}
\usepackage{etoolbox}
\usepackage{color}
\title{A Text-Image Fusion Method with Data Augmentation Capabilities for Referring Medical Image Segmentation}
\name{\begin{tabular}{c} 
    Shurong Chai$^{1}$, Rahul Kumar JAIN$^{2}$, Rui Xu$^{3}$, Shaocong Mo$^{4}$, Ruibo Hou$^{1}$,  Shiyu Teng$^{1}$, Jiaqing Liu$^{1}$ \\
    Lanfen Lin$^{4}$, Yen-Wei Chen$^{1,*}$\thanks{*Corresponding Author: Yen-Wei Chen (chen@is.ritsumei.ac.jp)}
\end{tabular}}

\address{
    $^1$College of Information Science and Engineering, Ritsumeikan University, Osaka, Japan\\
    $^2$Tiwaki Co., Ltd., Kusatsu, Japan \\
    $^3$School of Software, Dalian University of Technology, Dalian, China\\
    $^4$College of Computer Science and Technology, Zhejiang University, Hangzhou, China
}

\begin{document}
\maketitle

\begin{abstract}
Deep learning relies heavily on data augmentation to mitigate limited data, especially in medical imaging. Recent multimodal learning integrates text and images for segmentation, known as referring or text-guided image segmentation. However, common augmentations like rotation and flipping disrupt spatial alignment between image and text, weakening performance. To address this, we propose an early fusion framework that combines text and visual features before augmentation, preserving spatial consistency. We also design a lightweight generator that projects text embeddings into visual space, bridging semantic gaps. Visualization of generated pseudo-images shows accurate region localization. Our method is evaluated on three medical imaging tasks and four segmentation frameworks, achieving state-of-the-art results. Code is publicly available on GitHub: \url{https://github.com/11yxk/MedSeg_EarlyFusion}.
\end{abstract}

\section{Introduction}

Medical image segmentation supports clinical tasks such as disease detection, treatment planning, and surgery. Recently, multimodal approaches combining visual and text features have gained attention \cite{zhang2024vision}. Text incorporation enhances medical context understanding and boosts segmentation accuracy.

\begin{figure}[t]
\centering
\includegraphics[width=1\linewidth]{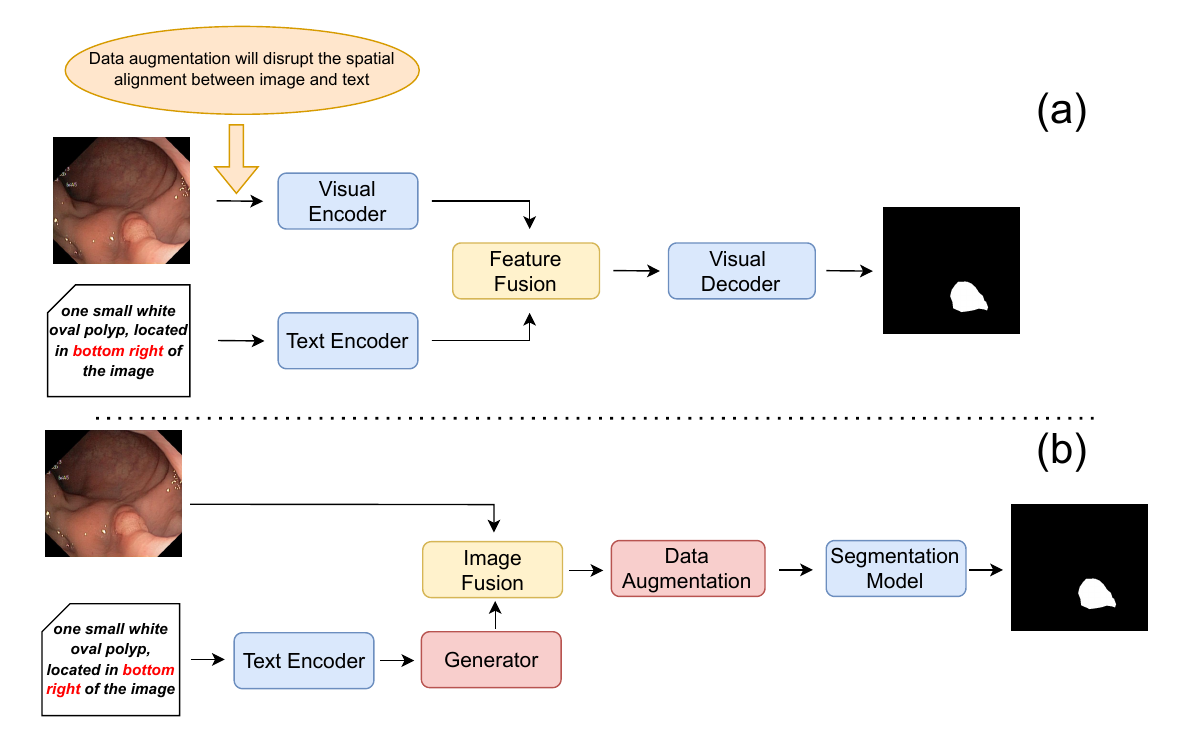}
\caption{Comparison between conventional method (a) and proposed method (b).}
\vspace{-15pt}
\label{fig:1}
\end{figure}

Data augmentation is also crucial in medical imaging \cite{chlap2021review}, especially with limited datasets. Due to costly, time-consuming annotations, acquiring medical images is difficult. Thus, augmentations like rotation, flipping, and scaling introduce variation, help deep learning models learn dataset patterns, and improve generalization and performance.

However, these data augmentation methods can significantly disrupt the alignment of text features. For example, if a lesion is positioned on the left side of an image, its text description will correspond to that position. However, after spatial data augmentation, such as, horizontal flipping, the lesion may shift to the right, while its text description remains unchanged. This misalignment can lead to a significant feature mismatch. On the other hand, completely disabling augmentations results in a substantial drop in accuracy.

\begin{figure*}[h]
\centering
\includegraphics[width=0.65\linewidth]{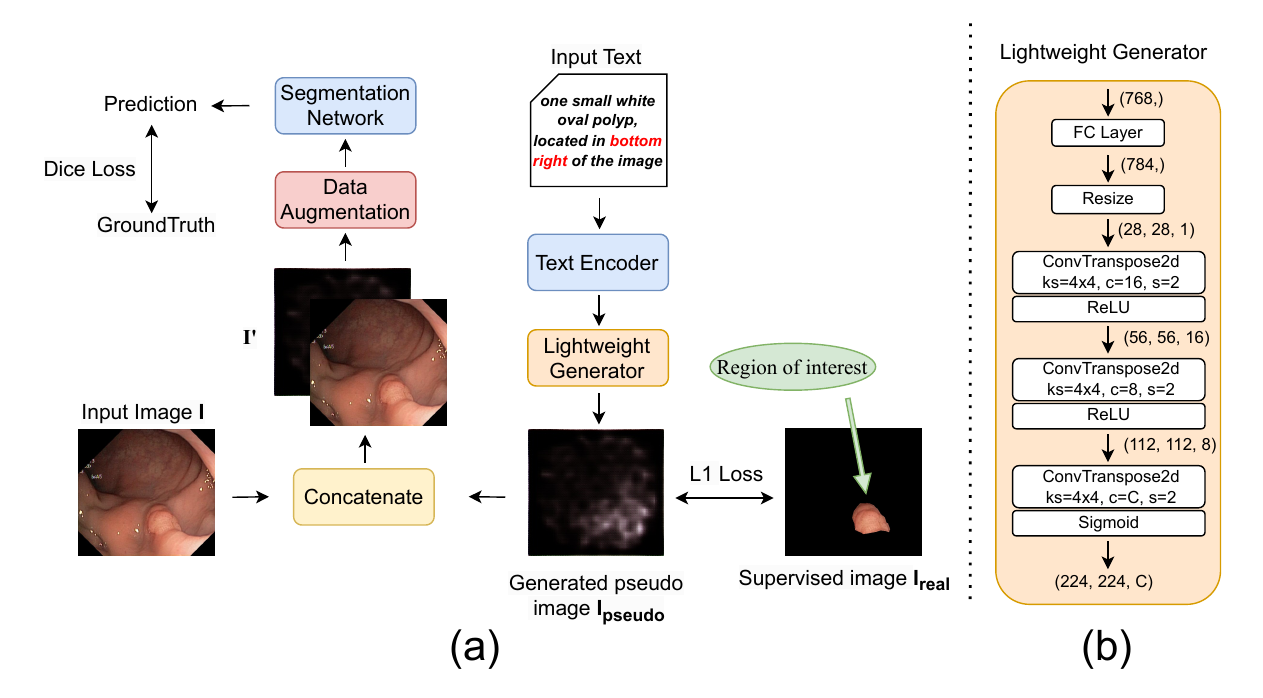}
\caption{Overview of the proposed early fusion approach. ks, c, s represent the kernel size, output channels and stride, respectively.}
\vspace{-15pt}
\label{fig:Overview}
\end{figure*}

In this study, we propose a novel early fusion approach that integrates text and visual information before data augmentation while ensuring spatial alignment between the image and its corresponding text description. We also introduce a generator based on a Region of Interest (ROI)-assisted learning to enhance the extraction of text embeddings from text descriptions. The comparison between the conventional text-guided method and the proposed method is shown in Figure~\ref{fig:1}. Our method maintains computational efficiency while being comparable to existing multimodal methods. This work highlights the importance of early fusion in medical image segmentation and opens new ways for providing practical solutions in real-world clinical applications. Our contributions can be summarized as follows:

1. We introduce a novel early fusion method for integrating text and visual information while enabling the application of data augmentation. This method effectively addresses the potential misalignment between text and image information, providing a practical approach to multimodal integration. 

2. We propose a text-generator-based scheme that learns to generate text features by utilizing the guidance from the ROI within the images. 

3. We provide comprehensive visualizations that illustrate the interaction between text and image features, demonstrating the alignment and effectiveness of our proposed approach in guiding segmentation tasks.

4. We validate our proposed methods on three different types of medical image datasets and four well-established segmentation frameworks. This extensive evaluation demonstrates the generalizability and robustness of our methods.

\section{Proposed Method}

\subsection{Overview}
Our proposed framework consists of a text encoder, a segmentation framework, and a lightweight generator for generating pseudo images. We use the combination of Dice loss and L1 loss to supervise the whole network. Figure~\ref{fig:Overview} illustrates the mechanism.

\subsection{Text-Generator-based Early Fusion}

\begin{figure}[t]
\centering
\includegraphics[width=1\linewidth]{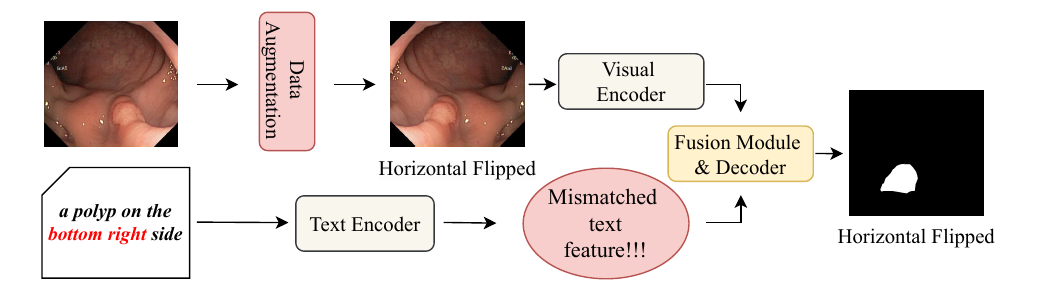}
\caption{Motivation of proposed method. Example illustrating the limitation of data augmentation in conventional fusion methods. When a horizontal flip is applied to the input image, the associated free-text description (e.g., “a polyp on the bottom right side”) becomes semantically inconsistent with the augmented image (the polyp is on the bottom left side).}
\vspace{-15pt}
\label{data_aug_moti}
\end{figure}

While spatial information is essential in computer vision tasks, most conventional multimodal approaches do not consider position-based data augmentation \cite{li2023lvit,ye2024enabling,zhang2024madapter,zhong2023ariadne}, resulting in suboptimal performance, especially when data is limited. As shown in Figure~\ref{data_aug_moti}, these data augmentation methods can significantly disrupt the alignment of text features. To address this fundamental issue, we propose an approach that projects text information for integration with images before providing it into the framework. This scheme enables the use of all types of data augmentation, which are crucial for model learning. Further, due to the differences between text features and image features, directly fusing them may bring noise to the original image. Therefore, to utilize the text information more effectively, we incorporate a lightweight generator to produce a more reliable “pseudo image”, mitigating the semantic gap between image and text modalities. The generated text feature map helps bridge the semantic gap between text and image modalities by utilizing the ROI as groundtruth (guidance signal).

\begin{table*}[t]
\centering
\caption{Experiment results with three datasets and four baseline methods. All the experiments were conducted five times to obtain the average and standard deviation. }  
\resizebox{0.5\textwidth}{!}{
\begin{tabular}{cc|cc|cc|cc}
\hline
\multicolumn{1}{c}{\multirow{2}{*}{Aug.}} & \multicolumn{1}{c|}{\multirow{2}{*}{Text}} & \multicolumn{2}{c|}{QATA-Covid}       & \multicolumn{2}{c|}{Kvasir}            & \multicolumn{2}{c}{ISIC 2016}             \\ \cline{3-8} 
\multicolumn{1}{c}{}   & \multicolumn{1}{c|}{}                       & \multicolumn{1}{c|}{Dice(\%)$\uparrow$}    & mIoU(\%)$\uparrow$   & \multicolumn{1}{c|}{Dice(\%)$\uparrow$}    & mIoU(\%)$\uparrow$    & \multicolumn{1}{c|}{Dice(\%)$\uparrow$}   & mIoU(\%)$\uparrow$   \\ \hline

\multicolumn{2}{c|}{} &\multicolumn{6}{c}{UNet} \\ \hline

$\mathbf{x}$       & $\mathbf{x}$      & \multicolumn{1}{c|}{86.17$\pm$0.3} & 75.71$\pm$0.5 & \multicolumn{1}{c|}{83.66$\pm$0.4} & 71.91$\pm$0.6 & \multicolumn{1}{c|}{90.35$\pm$0.1} & 82.41$\pm$0.2 \\
$\checkmark$        & $\mathbf{x}$    & \multicolumn{1}{c|}{87.43$\pm$0.1} & 77.67$\pm$0.2 & \multicolumn{1}{c|}{85.61$\pm$0.5} & 74.85$\pm$0.8 & \multicolumn{1}{c|}{91.46$\pm$0.1} & 84.27$\pm$0.2 \\
$\mathbf{x}$       & $\checkmark$   & \multicolumn{1}{c|}{89.64$\pm$0.1} & 81.22$\pm$0.3 & \multicolumn{1}{c|}{87.15$\pm$0.6} & 77.20$\pm$0.9 & \multicolumn{1}{c|}{91.63$\pm$0.2} & 84.56$\pm$0.4 \\
$\checkmark$        & $\checkmark$   & \multicolumn{1}{c|}{\textbf{90.46$\pm$0.1}} & \textbf{82.58$\pm$0.2} & \multicolumn{1}{c|}{\textbf{89.31$\pm$0.2}} & \textbf{80.69$\pm$0.4} & \multicolumn{1}{l|}{\textbf{92.78$\pm$0.1}} & \textbf{86.54$\pm$0.2} \\ \hline

\multicolumn{2}{c|}{} &\multicolumn{6}{c}{UNet++} \\ \hline
$\mathbf{x}$       & $\mathbf{x}$      & \multicolumn{1}{c|}{86.18$\pm$0.1} & 75.71$\pm$0.2 & \multicolumn{1}{c|}{85.23$\pm$1.0} & 74.28$\pm$1.6 & \multicolumn{1}{c|}{90.65$\pm$0.2} & 82.91$\pm$0.4 \\
$\checkmark$        & $\mathbf{x}$    & \multicolumn{1}{c|}{87.39$\pm$0.2} & 77.61$\pm$0.3 & \multicolumn{1}{c|}{86.83$\pm$0.2} & 76.73$\pm$0.3 & \multicolumn{1}{c|}{91.43$\pm$0.1} & 84.22$\pm$0.3 \\
$\mathbf{x}$      & $\checkmark$   & \multicolumn{1}{c|}{89.90$\pm$0.0} & 81.65$\pm$0.0 & \multicolumn{1}{c|}{87.66$\pm$0.4} & 78.03$\pm$0.6 & \multicolumn{1}{c|}{91.94$\pm$0.1} & 85.09$\pm$0.1 \\
$\checkmark$       & $\checkmark$   & \multicolumn{1}{c|}{\textbf{90.57$\pm$0.1}} & \textbf{82.77$\pm$0.1} & \multicolumn{1}{c|}{\textbf{89.14$\pm$0.4}} & \textbf{80.42$\pm$0.7} & \multicolumn{1}{c|}{\textbf{92.53$\pm$0.1}} & \textbf{86.10$\pm$0.1} \\ \hline

\multicolumn{2}{c|}{} &\multicolumn{6}{c}{TransUNet} \\ \hline
$\mathbf{x}$        & $\mathbf{x}$      & \multicolumn{1}{c|}{84.30$\pm$0.3} & 72.87$\pm$0.5 & \multicolumn{1}{c|}{80.91$\pm$1.9} & 67.98$\pm$2.6 & \multicolumn{1}{c|}{91.25$\pm$0.3} & 83.91$\pm$0.5 \\
$\checkmark$      & $\mathbf{x}$    & \multicolumn{1}{c|}{86.72$\pm$0.1} & 76.56$\pm$0.1 & \multicolumn{1}{c|}{86.85$\pm$0.3} & 76.77$\pm$0.5 & \multicolumn{1}{c|}{91.21$\pm$0.7} & 83.84$\pm$1.3 \\

$\mathbf{x}$     & $\checkmark$   & \multicolumn{1}{c|}{89.07$\pm$0.1} & 80.29$\pm$0.1 & \multicolumn{1}{c|}{82.66$\pm$0.9} & 70.46$\pm$1.3 & \multicolumn{1}{c|}{91.51$\pm$0.1} & 84.36$\pm$0.3 \\

$\checkmark$        & $\checkmark$   & \multicolumn{1}{c|}{\textbf{89.57$\pm$0.1}} & \textbf{81.16$\pm$0.1} & \multicolumn{1}{c|}{\textbf{87.17$\pm$0.5}} & \textbf{77.27$\pm$0.9} & \multicolumn{1}{c|}{\textbf{92.42$\pm$0.2}} & \textbf{85.92$\pm$0.4} \\ \hline

\multicolumn{2}{c|}{} &\multicolumn{6}{c}{MISSFormer} \\ \hline
$\mathbf{x}$        & $\mathbf{x}$      & \multicolumn{1}{c|}{83.45$\pm$0.2} & 71.61$\pm$0.3 & \multicolumn{1}{c|}{75.93$\pm$0.8} & 61.21$\pm$1.1 & \multicolumn{1}{c|}{90.38$\pm$0.3} & 82.45$\pm$0.5 \\
$\checkmark$       & $\mathbf{x}$    & \multicolumn{1}{c|}{85.34$\pm$0.1} & 74.42$\pm$0.2 & \multicolumn{1}{c|}{80.22$\pm$1.4} & 67.01$\pm$2.0 & \multicolumn{1}{c|}{91.34$\pm$0.1} & 84.07$\pm$0.1 \\

$\mathbf{x}$      & $\checkmark$   & \multicolumn{1}{c|}{88.64$\pm$0.1} & 79.60$\pm$0.1 & \multicolumn{1}{c|}{82.32$\pm$0.9} & 69.97$\pm$1.2 & \multicolumn{1}{c|}{91.23$\pm$0.3} & 83.87$\pm$0.5 \\

$\checkmark$       & $\checkmark$   & \multicolumn{1}{c|}{\textbf{89.17$\pm$0.1}} & \textbf{80.46$\pm$0.2} & \multicolumn{1}{c|}{\textbf{83.51$\pm$0.9}} & \textbf{71.70$\pm$1.3} & \multicolumn{1}{c|}{\textbf{92.24$\pm$0.2}} & \textbf{85.61$\pm$0.4} \\ \hline

\end{tabular}
}
\label{table:1}
\end{table*}

Following \cite{zhong2023ariadne}, we first incorporate a CXR-BERT \cite{boecking2022making} text encoder to extract the initial text features. These features are then generated to match the input image dimensions.  Let the 768-dimensional extracted text feature vector be denoted as $\mathbf{B} \in \mathbb{R}^{768}$. Then, we project this vector into a 784-dimensional space using a Fully Connected (FC) layer:

\begin{equation}
\mathbf{B'} = \text{FC}(\mathbf{B}) \in \mathbb{R}^{784}
\end{equation}

\begin{table}[h]
\centering
\caption{Ablation study of direct interpolation and the proposed ROI-based learning approach using a text generator. (Network: UNet)}
\setlength{\tabcolsep}{3pt} 
\scriptsize 
\resizebox{\columnwidth}{!}{
\begin{tabular}{c|cc|cc|cc}
\hline
\multirow{2}{*}{Method} & \multicolumn{2}{c|}{QATA-Covid} & \multicolumn{2}{c|}{Kvasir} & \multicolumn{2}{c}{ISIC 2016} \\ \cline{2-7}
& Dice(\%)$\uparrow$ & mIoU(\%)$\uparrow$ & Dice(\%)$\uparrow$ & mIoU(\%)$\uparrow$ & Dice(\%)$\uparrow$ & mIoU(\%)$\uparrow$ \\ \hline
Interpolation & 90.08 & 81.96 & 88.75 & 79.78 & 92.19 & 85.51 \\ \hline
Ours & \textbf{90.46} & \textbf{82.58} & \textbf{89.31} & \textbf{80.69} & \textbf{92.78} & \textbf{86.54} \\ \hline
\end{tabular}
}
\label{table:2}
\end{table}

\noindent Next, we reshape the projected feature $\mathbf{B'}$ into a $28 \times 28 \times 1$ matrix and then we use three stacked Transpose Convolution (TC) layers to generate a refined pseudo image $\mathbf{I_{\text{pseudo}}}$, where $H$, $W$ and $C$ are the height, width and channel of the input image:

\begin{equation}
\mathbf{B''} = \text{Resize}(\mathbf{B'}) \in \mathbb{R}^{28 \times 28 \times 1}
\end{equation}

\begin{equation}
\mathbf{I_{\text{pseudo}}} = \text{TC}(\mathbf{B''}) \in \mathbb{R}^{H \times W \times C}
\end{equation}

\noindent This generated pseudo image is then concatenated with the original input image $\mathbf{I} \in \mathbb{R}^{H \times W \times C}$:

\begin{equation}
\mathbf{I'} = \text{Concat}(\mathbf{I}, \mathbf{I_{\text{pseudo}}}) \in \mathbb{R}^{H \times W \times 2C}
\end{equation}

\noindent Since we use a robust generation mechanism guided by the ROI as groundtruth, we prefer to use the same number of channels (i.e., 2C) to provide richer multimodal information. We then apply data augmentation to the concatenated tensor using the advanced library Kornia \cite{eriba2019kornia}, making the training process both differentiable and practical.

\begin{equation}
\mathbf{I'_{\text{aug}}} = \text{DataAugmentation}(\mathbf{I'}) 
\end{equation}

\noindent Finally, the augmented tensor $\mathbf{I'_{\text{aug}}}$ is input into the network to generate a segmentation mask. In general, proposed method offers three key advantages over traditional methods: (a) It enables the network to perform any data augmentation. By applying various data augmentation techniques, the model can better handle text-image data, leading to improved generalization. (b) In comparison to recently proposed late fusion methods, which usually have a complex structure while our early fusion design is simple and effective. It achieves comparable performance and can be easily integrated into various existing methods, and requires only a slight change or increase in the input channel (as mentioned 2C). (c) The computational overhead introduced by our method is negligible, making it highly efficient and suitable for real-time applications in clinical settings.

\begin{table*}[t]
\centering
\caption{Comparison with the state-of-the-art methods in QATA-Covid Dataset. "Pretrained" indicates whether the model utilizes a pretrained visual encoder. The backbone segmentation network of ours is UNet++ \cite{zhou2019unet++}.}
\resizebox{0.5\textwidth}{!}{
\begin{tabular}{c|c|c|c|cc}
\hline
Method         & Publisher & Fusion Type                                                                          & \begin{tabular}[c]{@{}c@{}}Pretrained\end{tabular} & Dice(\%)$\uparrow$  & mIoU(\%)$\uparrow$  \\ \hline
LViT\cite{li2023lvit}                  & TMI'23    & \multirow{4}{*}{\begin{tabular}[c]{@{}c@{}}Intermediate\\ /Late Fusion\end{tabular}}                  &  $\mathbf{x}$                                              & 84.92 & 73.79 \\
SGSeg\cite{ye2024enabling}             & MICCAI'24 &                  &    $\checkmark$               & 87.40 & 77.80 \\
LanGuideMedSeg\cite{zhong2023ariadne}  & MICCAI'23 &                  &     $\checkmark$             & 89.78 & 81.45 \\

MAdapter\cite{zhang2024madapter}       & MICCAI'24 &                  &     $\checkmark$         & 90.22 & 82.16 \\ \hline
Ours (w/o Aug.) &  -   &      \multirow{2}{*}{\begin{tabular}[c]{@{}c@{}}Early Fusion\end{tabular}}               &      $\mathbf{x}$          & 89.90 & 81.65 \\
Ours (w/ Aug.)   & -         &                             &     $\mathbf{x}$                                           &\textbf{90.57} & \textbf{82.77} \\ \hline
\end{tabular}
}
\vspace{-10pt}
\label{table:3}
\end{table*}

\vspace{-5pt}
\subsection{Loss Function}

To supervise the generator, we use the real image, but only the segmentation region is used (i.e., ROI). Let the groundtruth segmentation mask be $\mathbf{M} \in \mathbb{R}^{H \times W}$, where $\mathbf{M}$ is a binary mask indicating the segmentation region. The supervision is applied using the L1 loss:

\begin{equation}
\mathcal{L}_{1} = \|\mathbf{I_{\text{pseudo}}} - \mathbf{I_{\text{real}}}\|_1 \quad \text{where} \quad \mathbf{I_{\text{real}}} = \mathbf{I} \odot \mathbf{M}
\end{equation}

\noindent here, $\mathbf{I_{\text{real}}}$ is the real image with only the segmentation region retained, and $\odot$ denotes the element-wise multiplication. The total loss function combines the Dice loss and L1 loss as follows:

\begin{equation}
\mathcal{L}_{\text{total}} = \mathcal{L}_{\text{Dice}} + \lambda \times \mathcal{L}_{1}
\end{equation}

\noindent The hyperparameter $\lambda$ is empirically determined and set to 0.1 based on extensive experimental validation.

\section{Experiment}

\subsection{Datasets}

We used three different types of publicly available medical image segmentation datasets: lung infection area segmentation (QATA-Covid \cite{degerli2022osegnet}), polyp segmentation (Kvasir \cite{jha2020kvasir}), and skin lesion segmentation (ISIC 2016 \cite{gutman2016skin}). The image samples in the training, validation, and test sets for these datasets are as follows:

\noindent QATA-Covid:  Training=5716; Validation=1429; Test=2113

\noindent Kvasir:    $\quad$ $\quad$ \ \ \ Training=900;   Validation=100; Test=100

\noindent ISIC 2016: \ \ \ \  \ \ Training=810;  Validation=90; Test=379

\begin{figure}[!htbp]
\centering
\includegraphics[width=0.9\linewidth]{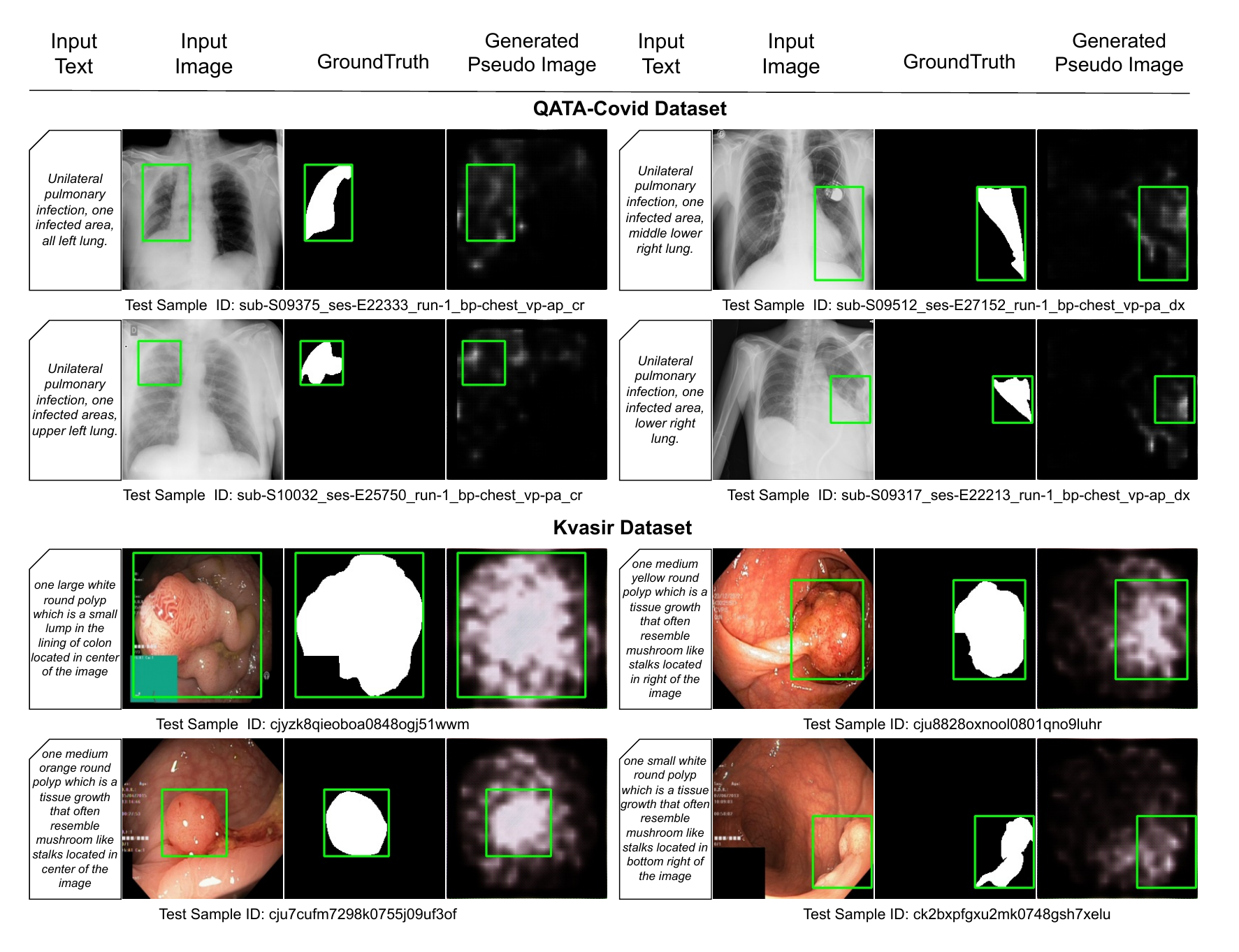}
\caption{Visualization results of generated pseudo images on the QATA-Covid and Kvasir datasets, demonstrating the alignment and effectiveness of our proposed approach.}
\vspace{-15pt}
\label{fig:3}
\end{figure}

\begin{figure}[!htbp]
\centering
\includegraphics[width=0.9\linewidth]{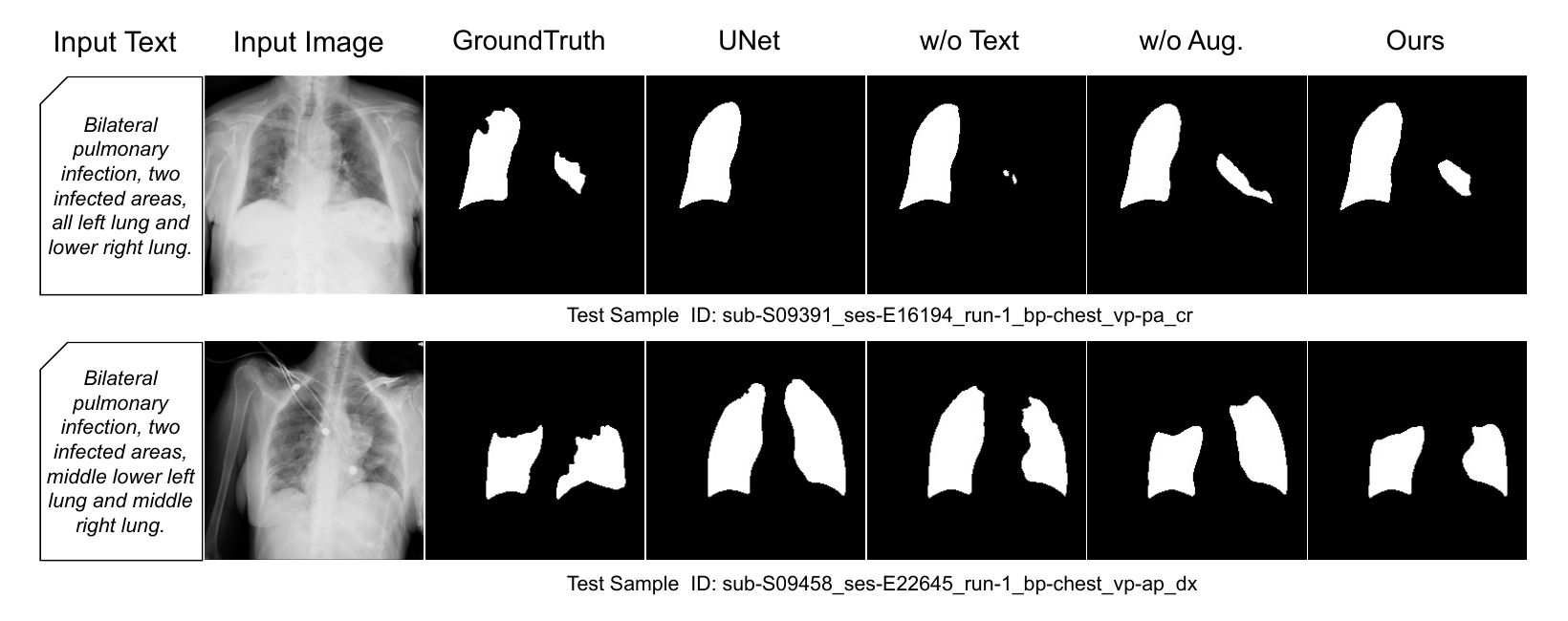}
\caption{Qualitative visualization results for the QATA-Covid dataset using UNet.}
\vspace{-15pt}
\label{fig:4}
\end{figure}

\subsection{Implementation Details}

The input image resolution is set to 224$\times$224. For training, we apply data augmentations, including random rotation, random translation, random scaling, and random flipping. The AdamW optimizer is used with a learning rate of 0.0001. We employ the CosineAnnealingLR scheduler from the PyTorch library to adjust the learning rate. All models are trained for 100 epochs with a batch size of 32. The text annotation for three datasets is from \cite{li2023lvit,poudel2023exploring}.

\vspace{-5pt}
\subsection{Experimental Results}
We evaluate our proposed method across four widely used frameworks. These include UNet \cite{ronneberger2015u} and UNet++ \cite{zhou2019unet++} (CNN-based), TransUNet \cite{chen2021transunet} (hybrid CNN and Transformer-based), and MISSFormer \cite{huang2022missformer} (Transformer-based).\par
The results using different frameworks and datasets are presented in Table~\ref{table:1}. Aug. represents the data augmentation. Across all datasets, our method significantly enhances the Dice score and mIoU compared to baseline. Table~\ref{table:2} presents the effectiveness of the proposed ROI-based learning approach using a text generator. As part of an ablation study, the extracted text features are projected and resized directly to match the input image dimensions by interpolation (without supervised L1 Loss and Transpose Convolution). The table provides results for both simple interpolated feature maps and the proposed ROI-based lightweight text generator (ours). Table~\ref{table:3} presents a comparison between our method and the SOTA approaches using the framework UNet++ \cite{zhou2019unet++}. Our method achieves competitive results without data augmentation and outperforms SOTA methods with augmentation, validating its effectiveness.

\vspace{-5pt}
\subsection{Visualization Results}
Figure~\ref{fig:3} presents the input text, image, groundtruth, and the generated pseudo image $\mathbf{I_{\text{pseudo}}}$. We can observe that the generated pseudo image highlights the segmentation areas and maintains the alignment between the text and the image, demonstrating the effectiveness of the lightweight generator’s guidance mechanism based on the image-based ROI. Figure~\ref{fig:4} presents qualitative visualization results for the QATA-Covid. Our method accurately segments infected regions as can be observed in the output.

\section{Conclusion and Future Work}

In this work, we primarily propose a method to utilize text-based multimodal information while enabling data augmentation techniques, addressing a crucial issue in recent research. In future work, we aim to explore more experiments involving other data augmentations, such as color transformations, CutMix and their alignment with text information.


\begin{thebibliography}{10}

\bibitem{zhang2024vision}
Jingyi Zhang, Jiaxing Huang, Sheng Jin, and Shijian Lu,
\newblock ``Vision-language models for vision tasks: A survey,''
\newblock {\em IEEE Transactions on Pattern Analysis and Machine Intelligence}, 2024.

\bibitem{chlap2021review}
Phillip Chlap, Hang Min, Nym Vandenberg, Jason Dowling, Lois Holloway, and Annette Haworth,
\newblock ``A review of medical image data augmentation techniques for deep learning applications,''
\newblock {\em Journal of medical imaging and radiation oncology}, vol. 65, no. 5, pp. 545--563, 2021.

\bibitem{li2023lvit}
Zihan Li, Yunxiang Li, Qingde Li, Puyang Wang, Dazhou Guo, Le~Lu, Dakai Jin, You Zhang, and Qingqi Hong,
\newblock ``Lvit: language meets vision transformer in medical image segmentation,''
\newblock {\em IEEE transactions on medical imaging}, vol. 43, no. 1, pp. 96--107, 2023.

\bibitem{ye2024enabling}
Shuchang Ye, Mingyuan Meng, Mingjian Li, Dagan Feng, and Jinman Kim,
\newblock ``Enabling text-free inference in language-guided segmentation of chest x-rays via self-guidance,''
\newblock in {\em International Conference on Medical Image Computing and Computer-Assisted Intervention}. Springer, 2024, pp. 242--252.

\bibitem{zhang2024madapter}
Xu~Zhang, Bo~Ni, Yang Yang, and Lefei Zhang,
\newblock ``Madapter: A better interaction between image and language for medical image segmentation,''
\newblock in {\em International Conference on Medical Image Computing and Computer-Assisted Intervention}. Springer, 2024, pp. 425--434.

\bibitem{zhong2023ariadne}
Yi~Zhong, Mengqiu Xu, Kongming Liang, Kaixin Chen, and Ming Wu,
\newblock ``Ariadne’s thread: Using text prompts to improve segmentation of infected areas from chest x-ray images,''
\newblock in {\em International Conference on Medical Image Computing and Computer-Assisted Intervention}. Springer, 2023, pp. 724--733.

\bibitem{boecking2022making}
Benedikt Boecking, Naoto Usuyama, Shruthi Bannur, Daniel~C Castro, Anton Schwaighofer, Stephanie Hyland, Maria Wetscherek, Tristan Naumann, Aditya Nori, Javier Alvarez-Valle, et~al.,
\newblock ``Making the most of text semantics to improve biomedical vision--language processing,''
\newblock in {\em European conference on computer vision}. Springer, 2022, pp. 1--21.

\bibitem{eriba2019kornia}
D.~Ponsa E.~Rublee E.~Riba, D.~Mishkin and G.~Bradski,
\newblock ``Kornia: an open source differentiable computer vision library for pytorch,''
\newblock in {\em Winter Conference on Applications of Computer Vision}, 2020.

\bibitem{zhou2019unet++}
Zongwei Zhou, Md~Mahfuzur~Rahman Siddiquee, Nima Tajbakhsh, and Jianming Liang,
\newblock ``Unet++: Redesigning skip connections to exploit multiscale features in image segmentation,''
\newblock {\em IEEE transactions on medical imaging}, vol. 39, no. 6, pp. 1856--1867, 2019.

\bibitem{degerli2022osegnet}
Aysen Degerli, Serkan Kiranyaz, Muhammad~EH Chowdhury, and Moncef Gabbouj,
\newblock ``Osegnet: Operational segmentation network for covid-19 detection using chest x-ray images,''
\newblock in {\em 2022 IEEE International Conference on Image Processing (ICIP)}. IEEE, 2022, pp. 2306--2310.

\bibitem{jha2020kvasir}
Debesh Jha, Pia~H Smedsrud, Michael~A Riegler, P{\aa}l Halvorsen, Thomas De~Lange, Dag Johansen, and H{\aa}vard~D Johansen,
\newblock ``Kvasir-seg: A segmented polyp dataset,''
\newblock in {\em MultiMedia modeling: 26th international conference, MMM 2020, Daejeon, South Korea, January 5--8, 2020, proceedings, part II 26}. Springer, 2020, pp. 451--462.

\bibitem{gutman2016skin}
David Gutman, Noel~CF Codella, Emre Celebi, Brian Helba, Michael Marchetti, Nabin Mishra, and Allan Halpern,
\newblock ``Skin lesion analysis toward melanoma detection: A challenge at the international symposium on biomedical imaging (isbi) 2016, hosted by the international skin imaging collaboration (isic),''
\newblock {\em arXiv preprint arXiv:1605.01397}, 2016.

\bibitem{poudel2023exploring}
Kanchan Poudel, Manish Dhakal, Prasiddha Bhandari, Rabin Adhikari, Safal Thapaliya, and Bishesh Khanal,
\newblock ``Exploring transfer learning in medical image segmentation using vision-language models,''
\newblock {\em arXiv preprint arXiv:2308.07706}, 2023.

\bibitem{ronneberger2015u}
Olaf Ronneberger, Philipp Fischer, and Thomas Brox,
\newblock ``U-net: Convolutional networks for biomedical image segmentation,''
\newblock in {\em Medical Image Computing and Computer-Assisted Intervention--MICCAI 2015: 18th International Conference, Munich, Germany, October 5-9, 2015, Proceedings, Part III 18}. Springer, 2015, pp. 234--241.

\bibitem{chen2021transunet}
Jieneng Chen, Yongyi Lu, Qihang Yu, Xiangde Luo, Ehsan Adeli, Yan Wang, Le~Lu, Alan~L Yuille, and Yuyin Zhou,
\newblock ``Transunet: Transformers make strong encoders for medical image segmentation,''
\newblock {\em arXiv preprint arXiv:2102.04306}, 2021.

\bibitem{huang2022missformer}
Xiaohong Huang, Zhifang Deng, Dandan Li, Xueguang Yuan, and Ying Fu,
\newblock ``Missformer: An effective transformer for 2d medical image segmentation,''
\newblock {\em IEEE transactions on medical imaging}, vol. 42, no. 5, pp. 1484--1494, 2022.

\end{thebibliography}
\end{document}